\documentclass{article}
\usepackage{spconf,amsmath,graphicx,amsfonts}
\usepackage{subcaption}
\usepackage{booktabs}
\usepackage{url}
\usepackage{fancyhdr}

\title{End-To-End Visual Speech Recognition With LSTMs}
\twoauthors
  { Stavros Petridis, Zuwei Li}
{Imperial College London\\
Dept. of Computing, London, UK\\
\{sp104;zl4615\}@imperial.ac.uk}
{Maja Pantic }
	{Imperial College London / Univ. of Twente\\
	Dept. of Computing, UK / EEMCS, Netherlands\\
	m.pantic@imperial.ac.uk}
\begin{document}
%\ninept
%
\maketitle
\begin{abstract}

Traditional visual speech recognition systems consist of two stages, feature extraction and classification. Recently, several deep learning approaches have been presented which automatically extract features from the mouth images and aim to replace the feature extraction stage. However, research on joint learning of features and classification is very limited. In this work, we present an end-to-end visual speech recognition system based on Long-Short Memory (LSTM) networks. To the best of our knowledge, this is the first model which simultaneously learns to extract features directly from the pixels and perform classification and also achieves state-of-the-art performance in visual speech classification.  The model consists of two streams which extract features directly from the mouth and difference images, respectively. The temporal dynamics in each stream are modelled by an LSTM and the fusion of the two streams takes place via a Bidirectional LSTM (BLSTM). 
 An absolute improvement of 9.7\% over the base line is reported on the OuluVS2 database, and 1.5\% on the CUAVE database when compared with other methods which use a similar visual front-end. 
\end{abstract}
\section{Introduction}
\label{sec:intro}

Speech is an audiovisual signal which consists of the audio vocalisation and the corresponding mouth configuration. Although most of the information is carried by the audio signal, the visual signal also carries  complementary and redundant information. 
This visual information, which is not affected by acoustic noise, can significantly improve the performance of speech recognition in noisy environments. 

Traditionally,  visual speech recognition systems consist of two stages, feature extraction from the mouth region of interest (ROI) and classification \cite{Potamianos2003, Dupont2000, Zhou2011}. 
The most common feature extraction approach is the use of a dimensionality reduction/compression method, with the most popular being the Discrete Cosine Transform (DCT), which results in a compact representation of the mouth ROI. In the second stage, a dynamic classifier, like Hidden Markov Models (HMMs) or Long-Short Term Memory (LSTM) recurrent neural networks, is used to model the temporal evolution of the features. 

Recently, several deep learning approaches for visual speech recognition have been presented. The vast majority also follow a two stage approach where deep bottleneck architectures are used for feature extraction. First, high dimensional features are extracted from the mouth ROI which are compressed to a low dimensional representation at the bottleneck layer of a deep network and then fed to a classifier.  Ngiam et al. \cite{ngiam2011multimodal} applied principal component analysis (PCA) to the mouth ROI and trained a deep autoencoder to extract bottleneck features. The features from the entire utterance were fed to a support vector machine ignoring the temporal dynamics of the speech.  Ninomiya et al. \cite{ninomiya2015integration} also applied PCA to the mouth ROIs and used a deep autoencoder to extract bottleneck features but an HMM was used in order to take into account the temporal dynamics. Sui et al. \cite{sui2014extracting} extracted local binary patterns from the mouth ROI and used a deep autoencoder to reduce their dimensionality. Then, the bottleneck features were concatenated with DCT features and fed to an HMM. A similar approach has also been followed in audiovisual speech recognition \cite{huang2013audio, mroueh2015, sui2015listening} where a shared representation of the input audio and visual features is extracted from the bottleneck layer.

\begin{figure}[t]

  \centering
\includegraphics[width=0.8\linewidth]{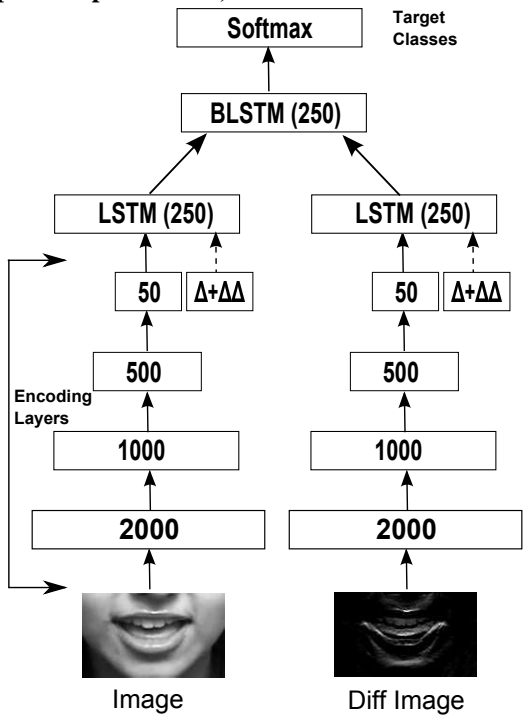}

\caption{Overview of the end-to-end visual speech recognition system. Two streams are used for feature extraction directly from the raw images. The first stream extracts features from the raw mouth ROI and the second stream from the diff mouth ROI in order to capture local temporal dynamics. The $\Delta$ and $\Delta\Delta$ features are also computed  and appended to the bottleneck layer. The encoding layers are pre-trained using RBMs. The temporal dynamics are modelled by an LSTM in each stream. A BLSTM is used to fuse the information from both streams and provides a label for each input frame. }
\label{fig:system}
\end{figure}

Few works have also been presented which extract bottleneck features directly from the pixels. Li \cite{Li2016} used a convolutional neural network (CNN) in order to extract bottleneck features from dynamic representations of images, which are fed to an HMM for classification.
In our previous work \cite{petridis2016deep}, we extracted bottleneck features directly from the raw mouth ROI using a deep feedforward network and then trained an LSTM for classification.  Noda et al. \cite{noda2015audio} used a CNN to predict the phoneme that corresponds to an input mouth ROI, and then an HMM is used together with audio features in order to classify an utterance.

Despite the success of deep learning methods in feature extraction, work on end-to-end visual speech recognition has been very limited. To the best of our knowledge, only Wand et al. \cite{wand2016lipreading} developed an end-to-end system for lipreading. The system consists of one feedforward layer followed by two LSTM layers and trained to perform lipreading directly from raw mouth ROIs. The system was tested on a subject-dependent experiment on the GRID corpus\cite{cooke2006} and although it outperformed other baseline systems it failed to outperform the state-of-the-art results \cite{Abdelaziz2015}. 

\looseness - 1
In this paper, we present an end-to-end visual speech recognition system which jointly learns the feature extraction and classification stages.  To the best of our knowledge, this is the first end-to-end model which performs visual speech recognition from raw mouth ROIs and achieves state-of-the-art performance. The system consists of two streams, one which encodes static information and one which encodes local temporal dynamics. The former operates on the raw mouth ROIs and the latter on the difference (diff) images. An LSTM models the temporal dynamics in each stream and the fusion of both streams occurs through a BLSTM. 

We perform subject independent experiments on two different datasets, OuluVS2 and CUAVE. An absolute improvement of 9.7\% over the baseline is reported on the OuluVS2 database, and 1.5\% on the CUAVE database when compared with other methods which use a similar visual front-end.

\begin{figure}[t]

  \centering
\includegraphics[width=0.5\linewidth]{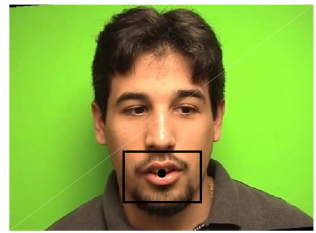}

\caption{Example of mouth ROI extraction from CUAVE}
\label{fig:mouthROIcuave}
\end{figure}

\section{Databases}
\label{sec:Databases}

\looseness - 1
The databases used in this study are the OuluVS2 \cite{Anina2015} and CUAVE \cite{Patterson:2002}. The OuluVS2 contains 52 speakers saying 10 utterances, 3 times each, so in total there are 156 examples per utterance. The utterances are the following: ``Excuse me", ``Goodbye", ``Hello", ``How are you", ``Nice to meet you", ``See you", ``I am sorry", ``Thank you", ``Have a good time", ``You are welcome".  The mouth ROIs are provided and they are downscaled to 26 by 44 in order to keep the aspect ratio constant.

\looseness - 1
The CUAVE dataset contains 36 subjects speaking digits 0 to 9, 5 times each, so in total there are 180 examples per digit. The normal portion of the database is used where the subjects are in frontal position. Sixty eight points are tracked on the face using the tracker proposed in \cite{Kazemi_2014_CVPR}. The faces are first aligned using a neutral reference frame in order to normalise them for rotation and size differences. This is done using an affine transform using 5 stable points, two eyes corners in each eye and the tip of the nose. Then the center of the mouth is located based on the tracked points and a bounding box
with size 90 by 150 is used to extract the mouth ROI as shown in Fig. \ref{fig:mouthROIcuave}. Finally, the mouth ROIs are downscaled to 30 by 50.

\section{End-To-End Visual Speech Recognition}
\label{sec:End2End}
The proposed deep learning system for visual speech recognition is shown  in Fig. \ref{fig:system}. It consists of two independent streams which extract features directly from the raw input. The first stream mainly encodes static information by extracting features directly from the raw mouth ROI. The second stream encodes the local temporal dynamics by extracting features from the diff mouth ROI, which is computed by taking the difference between two consecutive frames. 

Both streams follow a bottleneck architecture in order to compress the high dimensional input image to a low dimensional representation at the bottleneck layer.  The same architecture as in \cite{hinton2006reducing} is used, where 3 sigmoid hidden layers are used with sizes of 2000, 1000 and 500, respectively, followed by a linear bottleneck layer. These encoding layers are pre-trained in a greedy layer-wise manner using Restricted Boltzmann Machines (RBMs) \cite{hinton2012practical}. The $\Delta$ (first derivatives) and $\Delta\Delta$ (second derivatives) \cite{young2002htk} features are also computed, based on the bottleneck features, and they are appended to the bottleneck layer. In this way, during training we force the encoding layers to learn representations which produce good $\Delta$ and $\Delta\Delta$ features.

Finally, an LSTM layer is added on top of the encoding layers in order to model the temporal dynamics of the features in each stream. The LSTM outputs of each stream are concatenated and fed to a BLSTM in order to fuse the information from both streams. The output layer is a softmax layer which provides a label for each input frame. The entire system is trained end-to-end which enables the joint learning of features and classifier. In other words, the encoding layers learn to extract features from raw images which are useful for classification using LSTMs.

\begin{table}[t]
\renewcommand{\tabcolsep}{7pt}
\caption{Classification Accuracy on the OuluVS2 database. The end-to-end models are evaluated using the protocol suggested in \cite{ouluVS2} where 40 subjects are used for training and validation and 12 subjects are used for testing. $\dagger$ These models use a leave-one-subject-out cross validation for evaluation. }
\label{tab:resultsOuluVS}
\centering

\begin{tabular}{cc}
\toprule  Method & Classification    \\
           & Accuracy \\

\midrule End-to-End (Raw Image)   & 78.0 \\ 
\midrule End-to-End (Diff Image)   & 75.8  \\ 
\midrule End-to-End (Raw + Diff Images, Fig. \ref{fig:system})    & 84.5 \\

\midrule DCT + HMM \cite{ouluVS2} $\dagger$ & 74.8 \\
\midrule Latent Variable Models \cite{ouluVS2} $\dagger$   & 73.0  \\ 

\bottomrule
\vspace{-0.5cm}
\end{tabular} 

\end{table}

\begin{table}[t]
\renewcommand{\tabcolsep}{7pt}
\caption{Classification Accuracy on the CUAVE database. The end-to-end model is trained using the same protocol as \cite{ngiam2011multimodal, Srivastava:2014} where 18 subjects are used for training and validation and 18 for testing. $\ast$ This model is trained on 28 subjects and tested on 8 subjects. $\dagger$ These models are trained and tested using a 6-fold cross validation. $\ddagger$ This model uses a visual front-end which is significantly more complicated than ours.}
\label{tab:resultsCUAVE}
\centering

\begin{tabular}{cc}
\toprule Method  & Classification    \\
           & Accuracy \\
\midrule End-to-End (Raw Image)   & 71.4 \\ 
\midrule  End-to-End (Diff Image)   & 65.9  \\ 
\midrule  End-to-End (Raw + Diff Images, Fig. \ref{fig:system})    & 78.6 \\

\midrule   Deep Autoencoder + SVM \cite{ngiam2011multimodal} & 68.7 \\ 
\midrule Deep Boltzmann Machines + SVM \cite{Srivastava:2014} & 69.0 \\

\midrule AAM +HMM \cite{Papandreou2007} $\dagger$  & 75.7 \\
\midrule Patch-based Features + HMM \cite{Lucey:2006} $\ast$ & 77.1 \\
\midrule Visemic AAM + HMM \cite{Papandreou2009} $\dagger$ $\ddagger$ & 83.0 \\
\bottomrule
\vspace{-0.5cm}
\end{tabular} 

\end{table}

\section{EXPERIMENTAL SETUP}

\subsection{Evaluation Protocol}

We first partition the data into training and test sets. The protocol suggested by the creators of the OuluVS2 database is used \cite{ouluVS2} where 40 subjects are used for training and validation and 12 for testing. We randomly divided the 40 subjects into 30 and 10 subjects for training and validation purposes, respectively.  This means that there are 900 training utterances, 300 validation utterances and 360 test utterances. 

The evaluation protocol suggested in \cite{ngiam2011multimodal} was used for experiments on the CUAVE database. The odd-numbered subjects (18 in total) are used for testing and the even-numbered subjects are used for training. We further divided the latter ones into 12 subjects for training and 6 for validation. This means that there are 600, 300 and 900 training, validation and test utterances, respectively.

\subsection{Preprocessing}

Since all the experiments are subject independent we first need to reduce the impact of subject dependent characteristics. This is done by subtracting the mean image, computed over the entire utterance, from each frame.

\looseness - 1
The next step is the normalisation of data. As recommended in \cite{hinton2012practical} the data should be z-normalised, i.e. the mean and standard deviation should be equal to 0 and 1 respectively, before training an RBM with linear input units. Hence, each image is z-normalised before pre-training the encoding layers.

\subsection{Training}

\textbf{RBM Training:}  A Gaussian-Bernoulli RBM \cite{hinton2012practical} is used for the first layer since the input (pixels) is real-valued, followed by two Bernoulli-Bernoulli RBMs and one Bernoulli-Gaussian RBM for the linear bottleneck layer. Each RBM is trained for 20 epochs with a mini-batch size of 100 and L2 regularisation coefficient of 0.0002 using contrastive divergence. The learning rate is fixed to 0.1 for the Bernoulli-Bernoulli RBMs and to 0.001 when one layer (input or bottleneck) is real-valued as suggested in \cite{hinton2012practical}.

\noindent \looseness - 1
\textbf{End-to-End Training:} The AdaDelta algorithm \cite{Zeiler2012}, which automatically computes the learning rate in each epoch,  was used for training with a mini-batch size of 20 utterances. Early stopping with a delay of 5 epochs was also used in order to avoid overfitting. Gradient clipping was applied to the LSTM layers. The label of the last frame in each utterance was used in order to label the entire utterance.

\begin{figure}[t]

  \centering
\includegraphics[width=0.7\linewidth]{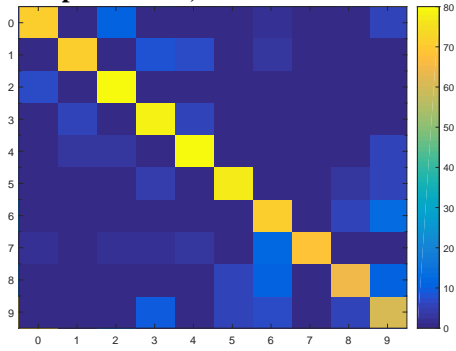}
  \caption{CUAVE confusion matrix. The labels for X and Y axes correspond to digits 0 to 9. }
\label{fig:confMatCUAVE}
\end{figure}

\section{RESULTS}
\label{sec:results}

\looseness - 1
Results for the OuluVS2 database are shown in Table \ref{tab:resultsOuluVS}. Since this database has been released recently only the baseline results provided by the creators are available. The best provided baseline result, 74.8\%, is achieved by HMMs in combination with DCT features. We first test each stream of the end-to-end model individually, i.e., just the encoding layers and the LSTM layer are considered. It is interesting to note that both streams outperform the baseline performance. The best overall result is achieved by the end-to-end 2-stream model, shown in Fig. \ref{fig:system}, with a classification accuracy of 84.5\%. We should also emphasise that the baseline performance is evaluated using a leave-one-subject-out cross validation approach which means there are 51 subjects for training and validation and only one subject for testing
in each iteration. On the other hand, we use much fewer subjects for training and validation, 40, and many more subjects for testing, 12, which makes the problem more challenging. Even in this case, the end-to-end system results in a significant improvement over the baseline performance.

Results for the CUAVE database are shown in Table \ref{tab:resultsCUAVE}. There is not a standard evaluation protocol for this database which makes comparison between different works difficult. Only \cite{ngiam2011multimodal} and \cite{Srivastava:2014} use the same evaluation protocol as in this study. We see that the single-stream end-to-end model based on raw mouth ROIs outperforms both previous works. The 2-stream end-to-end model outperforms all approaches that use a similar visual front-end. This includes \cite{Papandreou2007} where a 6-fold cross validation was used with 30 subjects for training and validation and 6 for testing, and \cite{Lucey:2006} where 28 subjects were used for training and validation and 10 for testing. In this study, we use much fewer subjects, 18, for training and validation and many more subjects for testing, 18. Only \cite{Papandreou2009} achieves a higher performance than our end-to-end system, but a much more complicated visual front-end is used, with a cascade of active appearance models (AAM),  and the model is evaluated using a 6-fold cross-validation.

\begin{figure}[t]
  \centering
 \includegraphics[width=0.7\linewidth]{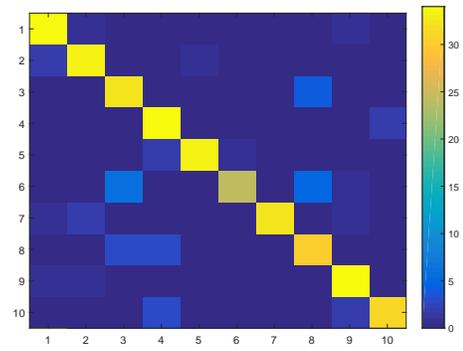}

  \caption{OuluVS2 confusion matrix. The labels for X and Y axes correspond to the 10 phrases described in section \ref{sec:Databases}.}
\label{fig:confMatOulu}
\end{figure}

Figures \ref{fig:confMatCUAVE} and \ref{fig:confMatOulu} show the confusion matrices for both datasets. In the OuluVS2 dataset, the most confusions were between phrases 3 (Hello) and 8 (Thank you) and between phrases 6 (See you) and 9 (Have a good time). In the CUAVE dataset, number pairs zero and two, six
and nine were most frequently confused. Zero and two share similar viseme
sequences near the end of the utterance while six and nine share similar viseme
sequences at the start of the utterance which explains the more frequently occurring
confusions for these number pairs.

Finally, we should also mention that we experimented with convolutional neural networks for the encoding layers but this led to worse performance than the proposed system. This is also reported in \cite{wand2016lipreading} and it is likely due to the small training sets. We also used data augmentation which improved the performance but did not exceed the performance of the proposed system.

\vspace{-0.4cm}

\section{CONCLUSION}
\looseness - 1
In this work, we present an end-to-end visual speech recognition system which 
jointly learns to extract features directly from the pixels and perform classification using LSTM networks. Results on subject independent experiments demonstrate that the proposed model achieves state-of-the-art performance on the OuluVS2 and CUAVE databases when compared with models which use a similar visual front end. The model can be easily extended to multiple streams so we are planning to add an audio stream in order to evaluate its performance on audiovisual speech recognition tasks.

\section{Acknowledgements}
This work has been funded by the European Community
Horizon 2020 under grant agreement no.
645094 (SEWA) and no. 688835 (DE-
ENIGMA).

%\vfill
%\pagebreak

%\vfill\pagebreak

\bibliographystyle{IEEEbib}
\bibliography{references}

\end{document}